\documentclass[onecolumn,letterpaper,11pt]{article}

\usepackage{xcolor, bbm, lastpage, color, amsfonts, cite, gensymb, here}
\usepackage{url, array, amsmath, fancyhdr, verbatim, graphicx, microtype}
\usepackage{amssymb, mathtools, bm}
\usepackage[footnotesize]{caption}
\usepackage[tmargin=1in,bmargin=1in,lmargin=1in,rmargin=1in]{geometry}
\usepackage{IEEEtrantools}
\usepackage[inline]{enumitem}
\usepackage{booktabs}
\usepackage{soul}
\usepackage{wrapfig}

\usepackage{titlesec}

\titleformat*{\section}{\large\bfseries}

\titleformat*{\subsection}{\normalsize\bfseries}
\usepackage{setspace}
\title{\vspace{-3cm} \singlespacing Research Report - Persistent Autonomy and Robot Learning Lab}
\author{S. Reza Ahmadzadeh \footnote{Persistent Autonomy and Robot Learning (PeARL) lab, Miner School of Computer and Information Sciences, University of Massachusetts Lowell, Lowell, 01854,  \texttt{reza\_ahmadzadeh@uml.edu}}}

\date{}
\begin{document}
\bstctlcite{IEEEexample:BSTcontrol}
%\pagenumbering{gobble}
\pagenumbering{arabic}
%%%%%%%%%%%%%%%%%%%%%%%

\maketitle
Robots capable of performing manipulation tasks in a broad range of missions in unstructured environments can develop numerous applications to impact and enhance human life~\cite{kormushev2015robot}. Existing work in robot learning has shown success in applying conventional machine learning algorithms to enable robots for replicating rather simple manipulation tasks in manufacturing~\cite{han2020towards, ahmadzadeh2017visuospatial}, service and healthcare applications, among others~\cite{ahmadzadeh2013interactive, ahmadzadeh2013visuospatial}. However, learning robust and versatile models for complex manipulation tasks that are inherently multi-faceted and naturally intricate demands algorithmic advancements in robot learning~\cite{han2020towards}.

Our research supports the long-term goal of making robots more accessible and serviceable to the general public by expanding robot applications to real-world scenarios that require systems capable of performing complex tasks. To achieve this goal, we focus on identifying and investigating knowledge gaps in robot learning of complex manipulation tasks by leveraging upon human-robot interaction and robot learning from human instructions.

This document presents an overview of the recent research developments in the Persistent Autonomy and Robot Learning (PeARL) lab at the University of Massachusetts Lowell. Here, I briefly discuss different research directions, and present a few proposed approaches in our most recent publications. For each proposed approach, I then mention potential future directions that can advance the field.

\section{Primitive Skill Modeling using Elastic Maps}

Several representations exist for modeling primitive skills from human demonstrations~\cite{rana2018towards, ahmadzadeh2017generalized, rana2017skill}. In this work, we proposed a novel optimization-based Learning from Demonstration (LfD) method that encodes demonstrations as elastic maps~\cite{hertel2022robot}. An elastic map is a graph of nodes connected through a mesh of springs. We build a skill model by fitting an elastic map to the set of demonstrations. The formulated optimization problem in our approach includes three objectives with natural and physical interpretations. The main term rewards the mean squared error in the Cartesian coordinate. The second term penalizes the non-equidistant distribution of points resulting in the optimum total length of the trajectory. The third term rewards smoothness while penalizing nonlinearity. These quadratic objectives form a convex problem that can be solved efficiently with local optimizers. We examined nine methods for constructing and weighting the elastic maps and study their performance in robotic tasks. We also evaluated the proposed method in several simulated and real-world experiments using a UR5e manipulator arm, and compare it to other LfD approaches to demonstrate its benefits and flexibility across a variety of metrics. In ~\cite{hertel2022robot}, we explain the proposed LfD representation using elastic maps and discuss our results compared against other methods.

Elastic maps present many opportunities to be combined with other LfD representations. Similar to other statistical representations, elastic maps could be used for generating a representative mean of demonstrated trajectories which then can be utilized by some other algorithms. The flexibility and configurability of elastic maps as well as their speed and ability to learn from a single demonstration allow them to be easily applied to various scenarios and show significant potential to be applied in modeling various robot skills for fast and accurate reproductions.

\begin{figure}[h]
    \centering
    \includegraphics[width=\textwidth]{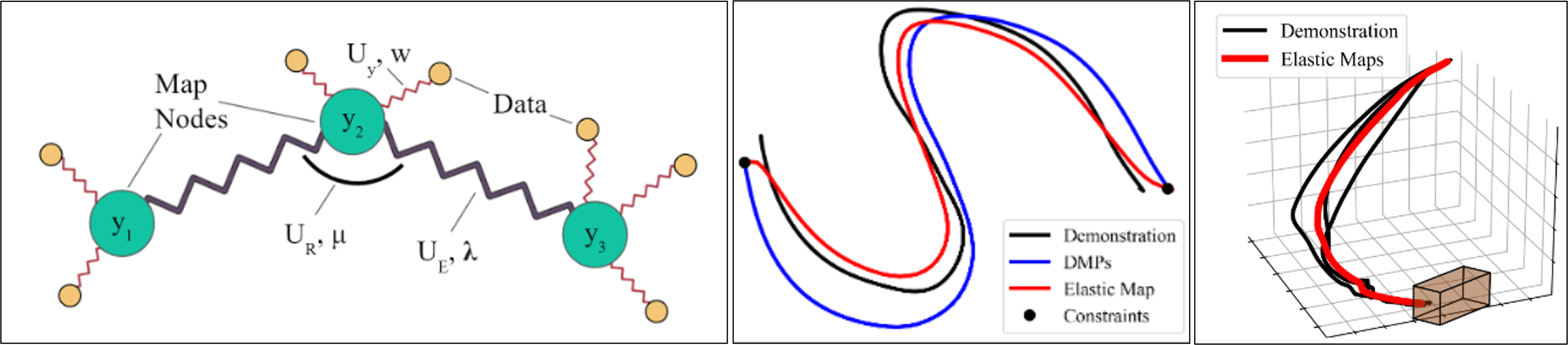}
    \caption{(left) Visualization of a simple polyline elastic map and associated energy elements, $U_y$, $U_E$, and $U_R$; (center) proposed approach compared against vanila DMPs in a 2D writing task;(right) reproduction of a pushing skill.}
    \label{fig:elasticmaps}
\end{figure}

\section{Constrained Skill Reproduction using Perturbation Analysis}

%Several methods exist for teaching robots, with one of the most prominent being Learning from Demonstration (LfD). 

Many LfD representations can be formulated as constrained optimization problems~\cite{ravichandar2019skill}. In this work, we proposed a novel convex formulation of the LfD problem~\cite{hertel2023confidence} represented as elastic maps~\cite{hertel2022robot}, which models reproductions as a series of connected springs. Relying on the properties of strong duality and perturbation analysis of the constrained optimization problem, we create a confidence metric. Our method allows the demonstrated skill to be reproduced with varying confidence level yielding different levels of smoothness and flexibility. Our confidence-based method provides reproductions of the skill that perform better for a given set of constraints. By analyzing the constraints, our method can also remove unnecessary constraints. We validated our approach using several simulated and real-world experiments using a Jaco2 7DOF manipulator arm. In ~\cite{hertel2023confidence}, we explain the proposed optimization formulation for learning constrained skills from demonstrations and discuss our results compared against other methods.

Beyond what we have shown here, there are several opportunities for future work. Firstly, we measure confidence by comparing a reproduction to the demonstration, assuming the demonstration to be the ``most confident.'' This may not always be the case because demonstrations may be noisy or otherwise sub-optimal. Instead, information from the surrounding environment could be used to find a better measure of confidence. Another avenue for future work is online adaptation of trajectories based on confidence. This paper considers stationary environments with trajectories computed a priori and executed with a low-level non-reactive controller. Real-world environments, however, are non-stationary and challenging, and require robots to adapt on-the-fly to the dynamic changes.

\begin{figure}[b]
    \centering
    \includegraphics[width=\textwidth]{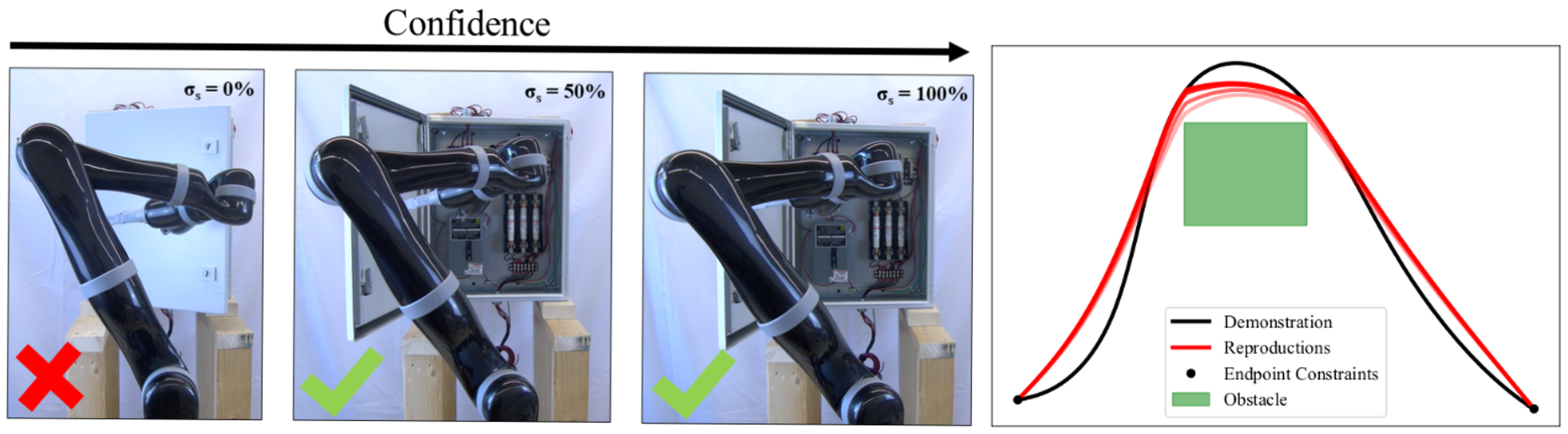}
    \caption{(left) Demonstration and reproductions with different confidence factors of opening a box. As confidence in a reproduction increases, the constraints tighten. The task cannot be reproduced successfully with a low level of confidence; (right) A 2D skill reproductions of varying levels of confidence, where confidence is shown with opacity.}
    \label{fig:confidence}
\end{figure}

\section{Skill Learning from Successful and Failed Demonstrations}

Learning from Demonstration (LfD) is a popular approach that allows humans to teach robots new skills by showing the \emph{correct} way(s) of performing the desired skill~\cite{ahmadzadeh2018trajectory}. Human-provided demonstrations, however, are not always optimal and the teacher usually addresses this issue by discarding or replacing sub-optimal (noisy or faulty) demonstrations~\cite{rana2020benchmark}. We proposed a novel LfD representation that learns from both successful and failed demonstrations of a skill~\cite{hertel2021learning}. Our approach encodes the two subsets of captured demonstrations (labeled by the teacher) into a statistical skill model, constructs a set of quadratic costs, and finds an optimal reproduction of the skill under novel problem conditions (i.e. constraints). The optimal reproduction balances convergence towards successful examples and divergence from failed examples. We evaluated our approach through several 2D and 3D experiments in real-world using a UR5e manipulator arm and also show that it can reproduce a skill from only failed demonstrations. The benefits of exploiting both failed and successful demonstrations were shown through comparison with two existing LfD approaches. We also compared our approach against an existing skill refinement method and show its capabilities in a multi-coordinate setting. In ~\cite{hertel2021learning}, we explain the details of the proposed method for learning from successful and failed demonstrations and discuss our evaluation results.

\begin{figure}[h]
    \centering
    \includegraphics[width=\textwidth]{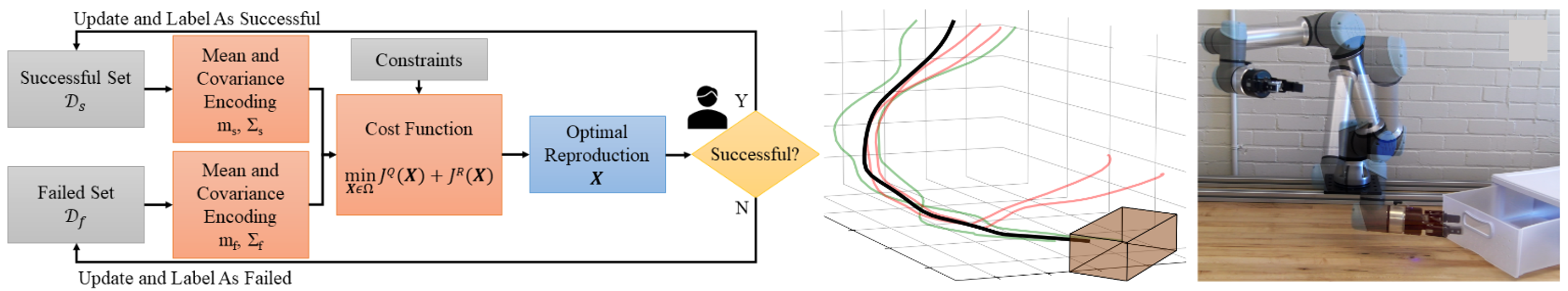}
    \caption{(left) workflow of the proposed approach; (center) and (right) a successful reproduction for a pushing task from successful and failed demonstrations with no constraints given.}
    \label{fig:tlsfd}
\end{figure}

Possible future work includes incorporating techniques such as skill refinement into the learning process, as well as improving the efficiency of the learning process. Alternatively, methods for improving the learning process could consider semantics, with the intent of understanding why a demonstration failed to better model failure and avoid reproducing it.

\section{Similarity-Aware Multi-Representational Skill Learning}

%Learning from Demonstration (LfD) algorithms
%enable humans to teach new skills to robots through demon-
%strations. 
Existing LfD methods can reproduce the modeled skills from the identical or near boundary conditions (e.g., initial point). However, when generalizing a learned skill over boundary conditions with higher variance, the similarity of the reproductions changes from one boundary condition to another, and a single LfD representation cannot preserve a consistent similarity across a generalization region. We proposed a novel similarity-aware framework including multiple LfD representations and a similarity metric that can improve skill generalization by finding reproductions with the highest similarity values for a given boundary condition~\cite{hertel2021similarity}. Given a demonstration of the skill, our framework constructs a similarity region around a point of interest (e.g., initial point) by evaluating individual LfD representations using the similarity metric. Any point within this volume corresponds to a representation that reproduces the skill with the greatest similarity. We validated our multi-representational framework in three simulated and four sets of real-world experiments using a physical 6-DOF robot. We also evaluated 11 different similarity metrics and categorize them according to their biases in 286 simulated experiments. In ~\cite{hertel2021similarity}, we explain the details of the proposed similarity-aware multi-representational framework and discuss our evaluation results.

One of the advantages of our framework is that it allows the user to select their metric of choice to evaluate and generalize the skill accordingly. While less experienced users can rely on a pre-selected similarity metric such as the Frechet distance, more experienced users can benefit from selecting a specific metric according to their task objectives.

\begin{figure}[h]
    \centering
    \includegraphics[width=\textwidth]{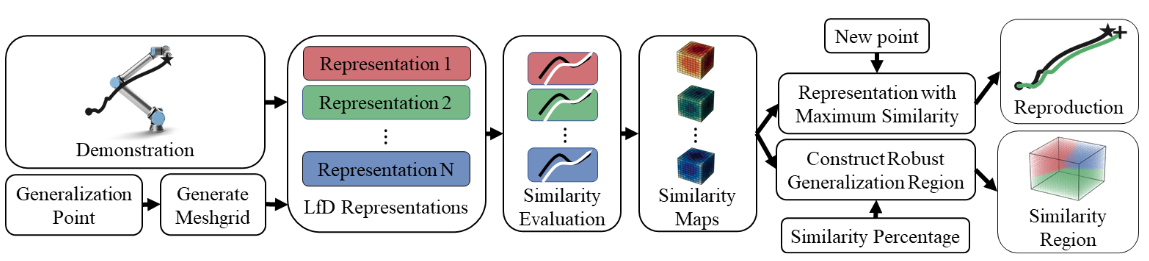}
    \caption{A flow diagram illustrating the proposed similarity-aware multi-representational skill learning framework.}
    \label{fig:simaware}
\end{figure}
%\input{teaching.tex}
%\clearpage
%\input{service.tex}
%\vspace{-1em}

%\pagenumbering{arabic}\setcounter{page}{1}
\bibliographystyle{IEEEtran}
\bibliography{references}

\end{document}